\documentclass[pdflatex]{nolta}
\usepackage[fleqn]{amsmath}
\usepackage{amsthm}
\usepackage{newtxtext}
\usepackage[varg]{newtxmath}
\usepackage{bm}
\usepackage{mathtools}
\usepackage{booktabs}
\usepackage{makecell}
\usepackage[english]{babel}

\newtheorem{proposition}{Proposition}

\Vol{2}%
\No{3}%
\Year{2011}%
\Month{10}%

\title{Self-organization and spectral mechanism of attractor
  landscapes in high-capacity kernel Hopfield networks}

\AUTHOR*{
\author{Akira Tamamori}{1}\orcid{0009-0000-8893-0058}
}

\AFFILIATE{ \affiliate{Faculty of Information Science, Aichi
    Institute of Technology\\1247 Yachigusa, Yakusa-cho, Toyota-shi, Aichi 470-0392, Japan}{1} }

\received{10}{27}{20XX}
\revised{12}{29}{20XX}
\published{7}{1}{20XX}

\begin{document}
\begin{abstract}
  Kernel-based learning methods can dramatically increase the storage
  capacity of Hopfield networks, yet the dynamical mechanisms behind
  this enhancement remain poorly understood. We address this gap by
  combining a geometric characterization of the attractor landscape
  with the spectral theory of kernel machines. Using a novel metric,
  \textit{Pinnacle Sharpness}, we empirically uncover a rich phase
  diagram of attractor stability, identifying a \textit{Ridge of
  Optimization} where the network achieves maximal robustness under
  high-load conditions. Phenomenologically, this ridge is characterized
  by a \textit{Force Antagonism}, in which a strong driving force is
  counterbalanced by a collective feedback force. We theoretically
  interpret this behavior as a consequence of a specific reorganization
  of the weight spectrum, which we term \textit{Spectral Concentration}.
  Unlike a simple rank-1 collapse, our analysis shows that the network
  on the ridge self-organizes into a critical regime: the leading
  eigenvalue is amplified to enhance global stability (Direct Force),
  while the trailing eigenvalues remain finite to sustain high memory
  capacity (Indirect Force). Together, these results suggest a spectral
  mechanism by which learning reconciles stability and capacity in
  high-dimensional associative memory models.
\end{abstract}

\maketitle

\section{Introduction}

Associative memory, the ability to retrieve full patterns from partial
or noisy cues, is a cornerstone of biological and artificial
intelligence. The Hopfield network~\cite{Hopfield1982} provides a
canonical model for this process, where memories are stored as stable
fixed points (attractors) of an energy landscape. However, the
classical model, governed by Hebbian learning, suffers from a severe
storage limit ($P \approx 0.14N$)~\cite{Amit1985}, restricting its
practical utility.

Recent years have seen a resurgence of interest in high-capacity
associative memories. Two dominant approaches have emerged. The first,
exemplified by \textit{Modern Hopfield Networks} (MHNs) and their
connection to Transformers~\cite{Krotov2016, Ramsauer2021}, achieves
capacity by redesigning the energy function itself (e.g., using
exponential interaction terms) to sharpen the attractors.  The second
approach, which is the focus of this study, retains the standard
quadratic energy structure (in a high-dimensional feature space) but
employs sophisticated \textit{learning algorithms} to optimize the
synaptic weights.  Our previous work demonstrated that Kernel Logistic
Regression (KLR) learning falls into this category, achieving storage
capacities ($P/N > 4.0$) comparable to MHNs while maintaining the
simplicity of kernel dynamics~\cite{tamamori2025b}.

Despite these performance gains, the underlying mechanism of
KLR-trained networks remains a black box. Why does a discriminative
learning rule (logistic regression) lead to a robust generative
landscape? Unlike Gardner's theory of capacity~\cite{Gardner1988a},
which focuses on the volume of solution space, we lack a dynamical
understanding of how the attractor geometry is sculpted. Specifically,
the geometric properties that enable stability under such high loads
are unknown.

This paper addresses this gap by unifying the
\textit{phenomenological} analysis of the attractor landscape with the
\textit{spectral} theory of kernel machines. We move beyond static
capacity metrics to investigate the local geometry of stability.

Our contributions are threefold:
\begin{enumerate}
\item \textbf{Phenomenological Discovery:} Using a novel metric,
  ``Pinnacle Sharpness,'' we uncover a rich phase diagram of attractor
  stability. We identify a ``Ridge of Optimization'' where the network
  achieves maximal robustness. On this ridge, we observe a ``Force
  Antagonism,'' where a strong driving force is balanced by a
  collective feedback force.
    
\item \textbf{Theoretical Mechanism:} We propose a spectral theory to
  explain these observations. We establish a link between the
  geometric stability and the leading eigenvalue of the kernel Gram
  matrix. Based on this link, we reveal that the Ridge corresponds to
  a state of \textbf{Spectral Concentration}. Unlike a simple rank-1
  collapse which would destroy memory capacity, our analysis shows
  that the network on the ridge self-organizes into a critical state:
  the leading eigenvalue is amplified to maximize global stability,
  while the trailing eigenvalues are preserved to maintain the rank
  required for multi-pattern storage.
    
\item \textbf{Unified Physical Picture:} We show that the observed
  Force Antagonism can be interpreted as a dynamical manifestation of
  Spectral Concentration. The dominant spectral mode creates a deep
  potential funnel (Direct Force), while the subordinate modes manage
  inter-pattern interference (Indirect Force).
\end{enumerate}
By elucidating this spectral mechanism, we provide a principled
explanation for the high performance of kernel associative memories,
distinguishing our learning-based approach from energy-based designs
like MHNs.

It is important to clarify the scope of our claims. In this work, we
use the term ``self-organization'' to refer to the empirically
observed and reproducible emergence of the Ridge of
Optimization. This usage should be distinguished from stronger
theoretical notions, such as self-organized criticality, which we
discuss later as a possible interpretation. The Ridge is therefore
presented as a structural regime identified through systematic
analysis, rather than as a mathematically guaranteed outcome of the
learning dynamics under all possible conditions.

The remainder of this paper is organized as follows.
Section~\ref{sec:related_work} reviews related work, positioning our
study within the context of capacity theory and kernel associative
memories.  Section~\ref{sec:methods} introduces the network model and
defines the geometric metrics used for our analysis.
Section~\ref{sec:results} presents the phenomenological results,
detailing the discovery of the Ridge and the associated Force
Antagonism.  Section~\ref{sec:theory} provides the theoretical core of
this work, deriving the spectral mechanism of stability and validating
the Spectral Concentration hypothesis.  Finally,
Section~\ref{sec:discussion} discusses the broader implications of our
findings, and Section~\ref{sec:conclusion} concludes the paper.

\section{Related Work}
\label{sec:related_work}

The quest to enhance the storage capacity and robustness of
associative memory has been a central theme in neural network research
for decades. Our work sits at the intersection of dynamical systems,
kernel methods, and modern dense associative memories.

\subsection{From Classical to Modern Hopfield Networks}
The classical Hopfield network~\cite{Hopfield1982}, governed by the
Hebbian learning rule, is theoretically elegant but practically
limited by a storage capacity of $P \approx 0.14N$~\cite{Amit1985}. To
overcome this, researchers have proposed various modifications to the
energy function.  Polynomial interaction terms were introduced to
increase capacity tensorially~\cite{Chen1986, Psaltis1986}. More
recently, Krotov and Hopfield proposed \textit{Dense Associative
  Memories} using rectified polynomial energy functions, significantly
boosting capacity~\cite{Krotov2016}. Demircigil et al. further
extended this to exponential energy functions, achieving exponential
storage capacity~\cite{Demircigil2017}.  A pivotal development was the
work by Ramsauer et al., which established the equivalence between
these exponential Hopfield networks and the attention mechanism in
Transformers~\cite{Ramsauer2021}.  These Modern Hopfield Networks
achieve high capacity by \textit{designing} steep energy landscapes
(e.g., Log-Sum-Exp). In contrast, our approach retains the standard
quadratic energy structure (in the feature space) but leverages a
powerful \textit{learning algorithm} (KLR) to sculpt the landscape
adaptively. This offers a complementary perspective: high capacity can
arise from learning dynamics rather than just architectural
definition.

\subsection{Learning Algorithms and Capacity Theory}
An alternative path to high capacity involves refining the learning
rule. Gardner's pioneering work established the theoretical capacity
limit for patterns to be linearly separable~\cite{Gardner1988a}. The
pseudo-inverse rule~\cite{Personnaz1986} and perceptron learning
provide mechanisms to approach this limit by uncorrelating patterns.

The introduction of kernel methods brought non-linear separation
capabilities to this domain. Early works extended associative memory
models to feature spaces, incorporating kernel functions directly into
the energy Hamiltonian~\cite{Caputo2002} or utilizing kernelized
pseudo-inverse rules~\cite{Nowicki2010}. Building on these
foundations, recent approaches have applied discriminative learning,
such as Support Vector Machines (SVMs) or Kernel Logistic Regression
(KLR), to train recurrent networks by interpreting the dynamics as an
iterative classification task. Our previous work established KLR as a
highly effective rule for this purpose, achieving state-of-the-art
capacity~\cite{tamamori2025, tamamori2025b}.

However, these studies typically evaluate performance based on static
classification metrics (e.g., margin) or final retrieval
capacity. They rarely investigate the \textit{dynamical mechanism};
specifically, it remains unclear how the learned weight spectrum
shapes the global attractor landscape during the transient recall
phase, which is the primary focus of this study.

\subsection{Geometry and Spectrum of Neural Dynamics}
Understanding the geometry of neural dynamics has been a longstanding
goal, often approached through statistical mechanics~\cite{Coolen2001}
or information geometry~\cite{Amari1999}. Amari and Wu pioneered the
geometric analysis of kernel machines, demonstrating how a kernel
function induces a Riemannian metric in the input space that governs
the spatial resolution of classification~\cite{Amari1999}.  Our work
complements these theoretical foundations by introducing a data-driven
spectral perspective. While statistical mechanics often assumes the
thermodynamic limit ($N \to \infty$) with random patterns, and
classical information geometry focuses on static classifiers, our
spectral analysis of the Ridge and Spectral Concentration provides a
concrete mechanism for the dynamics of finite-size networks. It
explains how the system self-organizes its weight spectrum to balance
stability (via the leading eigenvalue) and diversity (via the trailing
eigenvalues), offering a bridge between abstract geometric theories
and the actual stability of learned attractors.

\section{Model and Methods}
\label{sec:methods}

We consider a standard Hopfield network architecture trained using
Kernel Logistic Regression (KLR). In this section, we first define the
network model and the learning algorithm. We then introduce our
primary analytical tools: a geometric indicator for phenomenological
analysis and a spectral framework for theoretical interpretation.

\subsection{Kernel Logistic Regression Hopfield Network}
We consider a network of $N$ bipolar neurons,
$\bm{s} \in \{-1, 1\}^N$, trained to store $P$ patterns
$\{\boldsymbol{\xi}^\mu\}_{\mu=1}^P$. The KLR learning rule determines
the dual variables $\alpha_{\mu i}$ for each neuron $i$ by minimizing
a regularized logistic loss. For clarity, we collect these into a
matrix
$\bm{A} = [\boldsymbol{\alpha}_{1}, \ldots, \boldsymbol{\alpha}_{N}]
\in \mathbb{R}^{P\times N}$.  The network state $\bm{s}$ evolves
synchronously based on the input potential:
\begin{eqnarray}
  h_i(\bm{s}) &=& \sum_{\mu=1}^P \alpha_{\mu i} K(\bm{s}, \boldsymbol{\xi}^\mu) \\
  \quad s_i(t+1) &=& \mathrm{sign}(h_i(\bm{s}(t))),
\label{eq:update_rule}
\end{eqnarray}
where $K(\cdot, \cdot)$ is the RBF kernel with width parameter $\gamma$.

\subsection{Phenomenological Tool: Pinnacle Sharpness and Force Decomposition}
To quantitatively analyze the stability of the attractors, we
investigate the geometry of the network's energy landscape. While the
synchronous KLR dynamics do not guarantee the monotonic decrease of a
traditional energy function, we define a Lyapunov function candidate
whose landscape reflects the restorative forces of the system:
\begin{equation}
    V(\bm{s}) \coloneqq -\sum_{k=1}^N s_k h_k(\bm{s}),
    \label{eq:lyapunov_candidate}
\end{equation}
where $h_k(\bm{s})$ is the input potential defined in
Eq.~(\ref{eq:update_rule}). This function measures the alignment
between the current state $\bm{s}$ and the input field.

We utilize the ``Pinnacle Sharpness'', defined as the squared norm of
the gradient of this function at a stored pattern:
\begin{equation}
    M(\boldsymbol{\xi}^\mu) \coloneqq \|\nabla V(\bm{s})\|^2 \big|_{\bm{s}=\boldsymbol{\xi}^\mu}.
    \label{eq:pinnacle_sharpness}
\end{equation}
To calculate the gradient, we treat the discrete state $\bm{s}$ as a
continuous variable $\bm{x} \in \mathbb{R}^N$.  This continuous
relaxation is introduced solely as a local surrogate to probe the
stability geometry in the neighborhood of discrete fixed points,
rather than as an exact description of the underlying
dynamics. Differentiating Eq.~(\ref{eq:lyapunov_candidate}) yields the
gradient vector, which we analytically decompose into two functionally
distinct components:
\begin{equation}
    \nabla V(\bm{x}) = \bm{F}_d(\bm{x}) + \bm{F}_i(\bm{x}).
    \label{eq:force_decomposition}
\end{equation}
The first component, the \textbf{Direct Force} $\bm{F}_d$, is defined as:
\begin{equation}
    \bm{F}_d(\bm{x}) \coloneqq -\bm{h}(\bm{x}).
    \label{eq:direct_force}
\end{equation}
This term represents the immediate drive for the state vector to align
with its own input field. The second component, the \textbf{Indirect
  Force} $\bm{F}_i$, captures the collective feedback effects and
is given by:
\begin{equation}
    [\bm{F}_i(\bm{x})]_j = -\sum_{k=1}^N x_k \frac{\partial h_k(\bm{x})}{\partial x_j}.
    \label{eq:indirect_force}
\end{equation}
This term describes how a change in neuron $j$ perturbs the inputs to
all other neurons $k$, and how those perturbations reflexively
influence the energy. We measure the alignment between these two
forces using their cosine similarity $\rho(\bm{s})$.

\subsection{Theoretical Framework: Spectral Representation}
To provide a mathematical basis for the observed forces and stability,
we introduce a spectral representation of the network dynamics.

\textbf{Matrix Representation of Weights:} While kernel Hopfield
networks do not store an explicit $N \times N$ weight matrix, the
effective linear operator $\bm{W}$ in the high-dimensional feature
space $\mathcal{H}$ governs the dynamics. Crucially, this operator is
finite-rank and can be fully characterized by the stored patterns and
the learned dual variables.  Let $\phi: \mathbb{R}^N \to \mathcal{H}$
be the feature mapping associated with the RBF kernel, which maps the
input state into a high-dimensional Reproducing Kernel Hilbert Space
(RKHS) $\mathcal{H}$.  Let
$\boldsymbol{\Phi} = [\phi(\boldsymbol{\xi}^1), \dots,
\phi(\boldsymbol{\xi}^P)]$ be the feature matrix collecting the mapped
patterns.  While the network is trained by optimizing the per-neuron
dual variables $\alpha_{\mu i}$ (a $P \times N$ set of coefficients),
our theoretical analysis focuses on the collective interaction between
patterns in the feature space.  The weight operator $\bm{W}$ can be
represented as:
\begin{equation}
    \bm{W} = \boldsymbol{\Phi} \boldsymbol{\alpha} \boldsymbol{\Phi}^\top,
    \label{eq:weight_operator}
\end{equation}
where $\boldsymbol{\alpha} \in \mathbb{R}^{P \times P}$ is the
effective pattern interaction matrix.  For the purpose of our spectral
theory, we treat this $P \times P$ matrix $\boldsymbol{\alpha}$ as the
central object of analysis, representing an idealized model of the
interactions learned by the full set of $\alpha_{\mu i}$.  This matrix
can be understood as an effective representation of the pattern
correlations learned across all neurons, and can be approximated by
the Gram matrix of the per-neuron coefficients, i.e.,
$\boldsymbol{\alpha} \approx \frac{1}{N} \bm{A}\bm{A}^{\top}.$

\textbf{The Kernel Gram Matrix:} The geometry of the patterns in the
feature space is encoded in the kernel Gram matrix
$\bm{K} \in \mathbb{R}^{P \times P}$, with entries
$K_{\mu\nu} = K(\boldsymbol{\xi}^\mu, \boldsymbol{\xi}^\nu) = \langle
\phi(\boldsymbol{\xi}^\mu), \phi(\boldsymbol{\xi}^\nu) \rangle$.

\textbf{Spectral Connection:} A key insight of our theoretical
approach is that the stability of the dynamical system (the Hessian of
the energy) is spectrally equivalent to the product of the dual
coefficients and the Gram matrix. Specifically, the eigenvalues
governing the basin depth and shape are determined by the spectrum of
$\boldsymbol{\alpha}^{1/2} \bm{K} \boldsymbol{\alpha}^{1/2}$ (assuming
positive definite $\boldsymbol{\alpha}$).  This formulation allows us
to translate the physical ``sharpness'' of attractors directly into
the ``spectral gap'' of the kernel matrix, providing a bridge between
the learning algorithm (which optimizes $\boldsymbol{\alpha}$) and the
resulting physics of the memory landscape.

\textbf{Information Geometric Framework}: To analyze the learning
dynamics, we also employ tools from Information Geometry. The KLR
model forms a statistical manifold equipped with the Fisher
Information Matrix (FIM) as a Riemannian metric. For a single neuron,
the FIM $\bm{G}$ is given by $\bm{G}=\bm{K}\bm{D}\bm{K}$, where
$\bm{D}$ is a diagonal matrix of prediction variances. This metric
allows us to distinguish between the Euclidean Gradient $\nabla L$,
which guides standard SGD, and the Natural Gradient
$\tilde{\nabla}L = \bm{G}^{-1}\nabla L$ , which represents the
steepest descent direction on the manifold. The interplay between
these two gradients reveals the intrinsic geometry of the learning
process. A detailed derivation of these concepts is provided in
Appendix B.

\section{Phenomenology: Discovery of the Optimization Ridge}
\label{sec:results}
Our previous study established that KLR-trained networks achieve
exceptional storage capacity and noise
robustness~\cite{tamamori2025b}. Building on this performance
baseline, this section presents the results of large-scale numerical
simulations designed to uncover the dynamical mechanism governing this
stability.

We first demonstrate that the local stability of attractors,
quantified by our Pinnacle Sharpness metric, exhibits a complex
dependency on network parameters, revealing a rich phase structure. We
then dissect the underlying mechanism by analyzing the interplay
between the constituent forces of the landscape gradient.

\subsection{The Phase Diagram of Attractor Stability}
To obtain a global view of how attractor stability is organized, we
systematically computed the Pinnacle Sharpness, $M$, across a wide
range of the kernel locality parameter $\gamma$ and the storage load
$\beta = P/N$.  For each pair of $(\gamma, P/N)$, a KLR Hopfield
network with $N=100$ neurons was trained using the optimization
protocol detailed in~\cite{tamamori2025b}. The Pinnacle Sharpness was
then calculated at the center of a randomly chosen stored pattern
$\boldsymbol{\xi}^{\mu}$.  Throughout this paper, we refer to low-load
and high-load regimes according to whether $\beta$ is small or large
relative to the transition region observed in the phase diagram.

The results are summarized in the phase diagram shown in
Figure~\ref{fig:sharpness_diagram}. This figure plots the $\log_{10}$
of the Pinnacle Sharpness as a heatmap. The diagram reveals a
strikingly complex and organized structure, which can be characterized
by three distinct regions:

\begin{figure}[t]
\begin{center}
  \includegraphics[width=0.7\hsize]{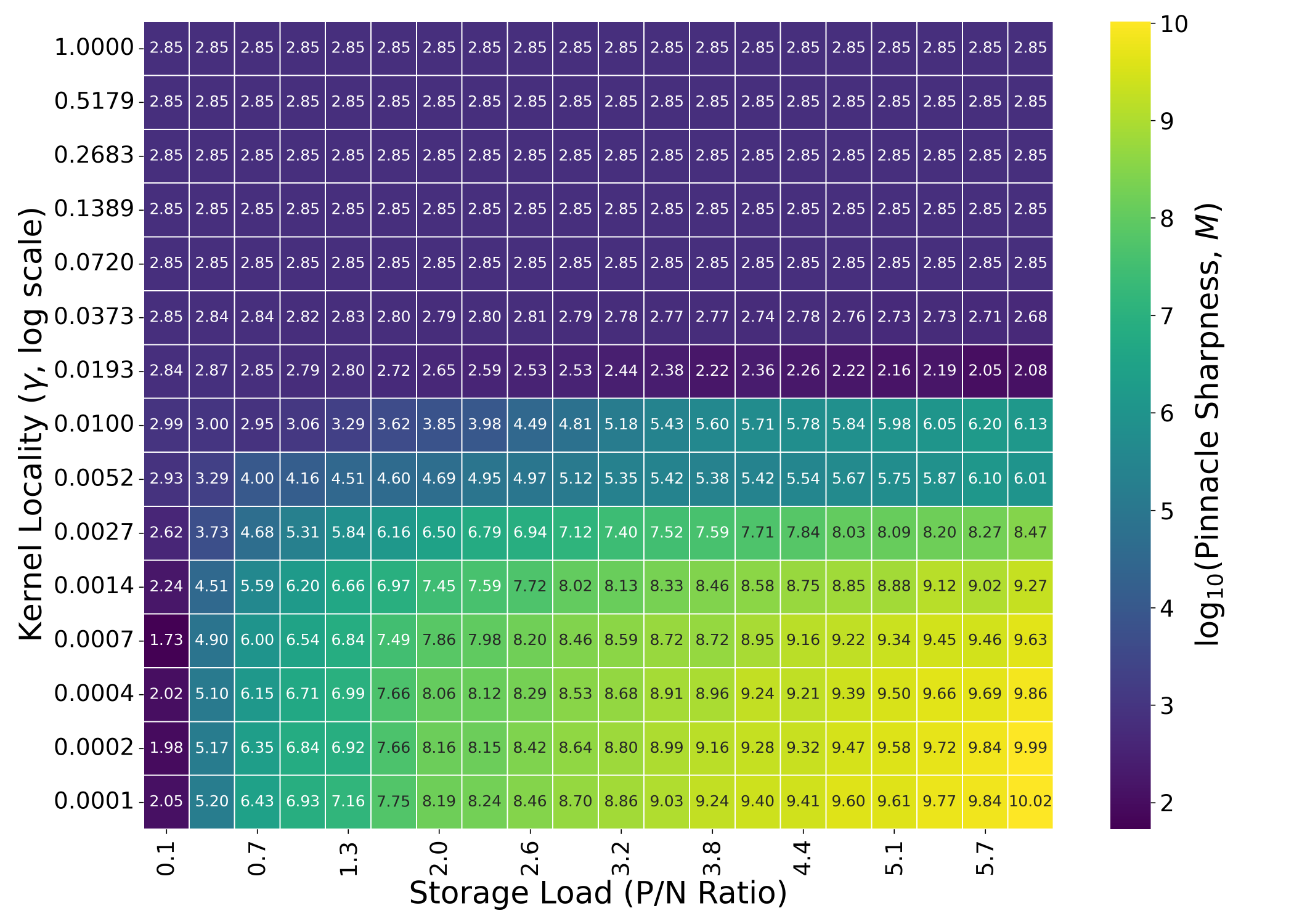}
  \caption{Phase diagram of attractor stability, quantified by the
    Pinnacle Sharpness $M$. The heatmap shows $\log_{10} M$ as a
    function of kernel locality $\gamma$ and storage load $P/N$. Three
    regimes are visible: (i) a local regime (large $\gamma$), (ii) a
    global, inefficient regime (small $\gamma$, low load), and (iii) a
    remarkable \textbf{Ridge of Optimization} (bright diagonal band)
    where the network achieves maximal stability. This ridge
    corresponds to the spectral ``Goldilocks zone'' identified in our
    theoretical analysis.}
  \label{fig:sharpness_diagram}
\end{center}
\end{figure}

\begin{enumerate}
\item \textbf{The Local, Non-Cooperative Regime}: In the upper region
  of the diagram, corresponding to large gamma values
  ($\gamma > 0.05$), the Pinnacle Sharpness remains largely constant
  at a moderate level, irrespective of the storage load $P/N$. In this
  regime, the kernel's influence is highly localized, suggesting that
  each attractor is stabilized independently without significant
  interaction from other stored patterns.
\item \textbf{The Global, Inefficient Regime}: In the lower-left
  region, characterized by small kernel locality $\gamma$ and low
  storage load $\beta$, the Pinnacle Sharpness drops to its lowest
  values. Here, the kernel is global, but the scarcity of patterns
  (low $\beta$) provides insufficient information for the learning
  algorithm to form sharp, well-defined attractors. The energy
  landscape is consequently flat and lacks strong restorative forces.
\item \textbf{The Ridge of Optimization}: The most remarkable feature
  of the phase diagram is the emergence of a ``ridge'' of extremely
  high Pinnacle Sharpness, extending diagonally from the lower-left to
  the lower-right. Along this ridge, the network achieves a level of
  attractor stability that is orders of magnitude greater than in the
  other regimes. This indicates that, faced with the dual challenges
  of a global kernel (small $\gamma$) and high memory congestion
  (large $P/N$), the network does not simply fail. Instead, it enters
  a highly optimized state, suggesting a sophisticated
  self-organization mechanism is at play.
\end{enumerate}

\subsection{Decomposing the Forces: The Mechanism of Self-Organization}
\begin{figure}[t]
\begin{center}
  \includegraphics[width=0.7\hsize]{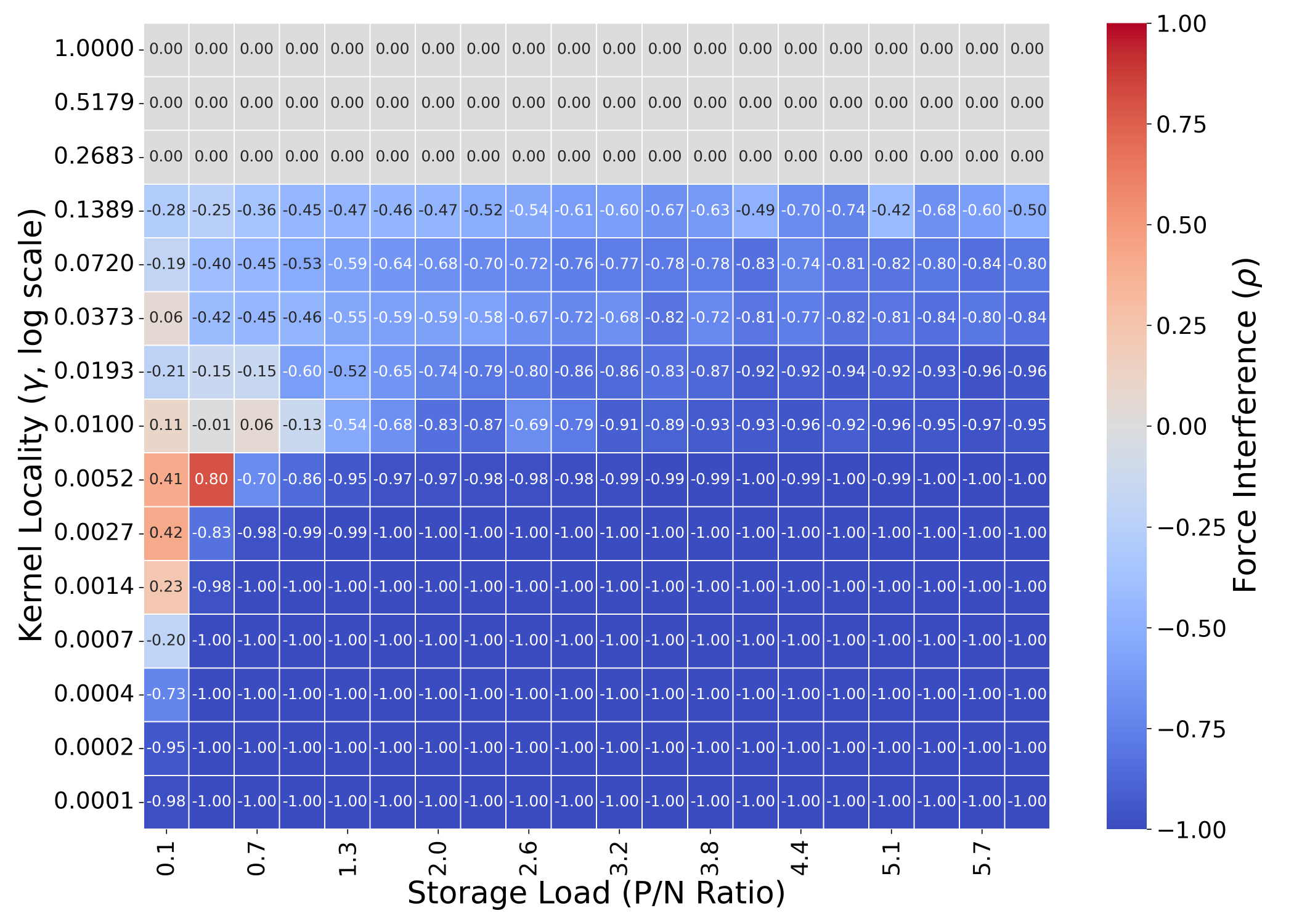}
  \caption{Phase diagram of Force Interference $\rho$, measuring the
    cosine similarity between the Direct Force $\bm{F}_d$ and
    Indirect Force $\bm{F}_i$. Blue regions indicate strong
    anti-correlation (antagonism, $\rho \approx -1$). The Ridge of
    Optimization in Fig.~\ref{fig:sharpness_diagram} precisely
    traces the transition boundary into this antagonistic regime,
    suggesting that stability is maintained by a delicate balance of
    opposing spectral forces.}
  \label{fig:interference_diagram}
\end{center}
\end{figure}

To understand the origin of the Ridge, we analyzed the behavior of the
two constituent forces of the landscape gradient: the direct force
$\bm{F}_d$ and the indirect feedback force $\bm{F}_i$. We computed the
normalized correlation between these two forces, the Force
Interference $\rho(\bm{x})$, across the same parameter space.

The resulting phase diagram is shown in
Figure~\ref{fig:interference_diagram}. This figure reveals a clear
transition. In the local regime (large $\gamma$), the interference is
approximately zero, confirming that the two forces act largely
independently. However, as gamma decreases, the interference rapidly
transitions to a state of strong anti-correlation ($\rho \approx
-1$). This indicates that in the global regime, the indirect feedback
force $\bm{F}_i$ actively opposes the direct force $\bm{F}_d$.

This strong anti-correlation provides the key to understanding the
ridge. The Pinnacle Sharpness, $\|\nabla V\|^2$, is given by
$\|\bm{F}_d\|^2 + \|\bm{F}_i\|^2 + 2\left<\bm{F}_d,
  \bm{F}_i\right>$. When $\rho \approx -1$, this simplifies to
$\|\nabla V\|^2 \approx (\|\bm{F}_d\| - \|\bm{F}_i\|)^2$. This means
that the Pinnacle Sharpness is determined by the difference in the
magnitudes of the two opposing forces.

\subsection{The Nature of the Optimization Ridge}
Finally, to elucidate the nature of the Ridge, we analyzed the
magnitudes of the individual forces along a representative
cross-section of the phase diagram. We fixed gamma to a small value
($\gamma = 0.001$), where the cooperative anti-correlation is strong,
and plotted the log-magnitudes of $\|\bm{F}_d\|^2$, $\|\bm{F}_i\|^2$,
and the resulting Pinnacle Sharpness ($\|\nabla V\|^2$) as a function
of the storage load $P/N$.

\begin{figure}[t]
\begin{center}
  \includegraphics[width=0.6\hsize]{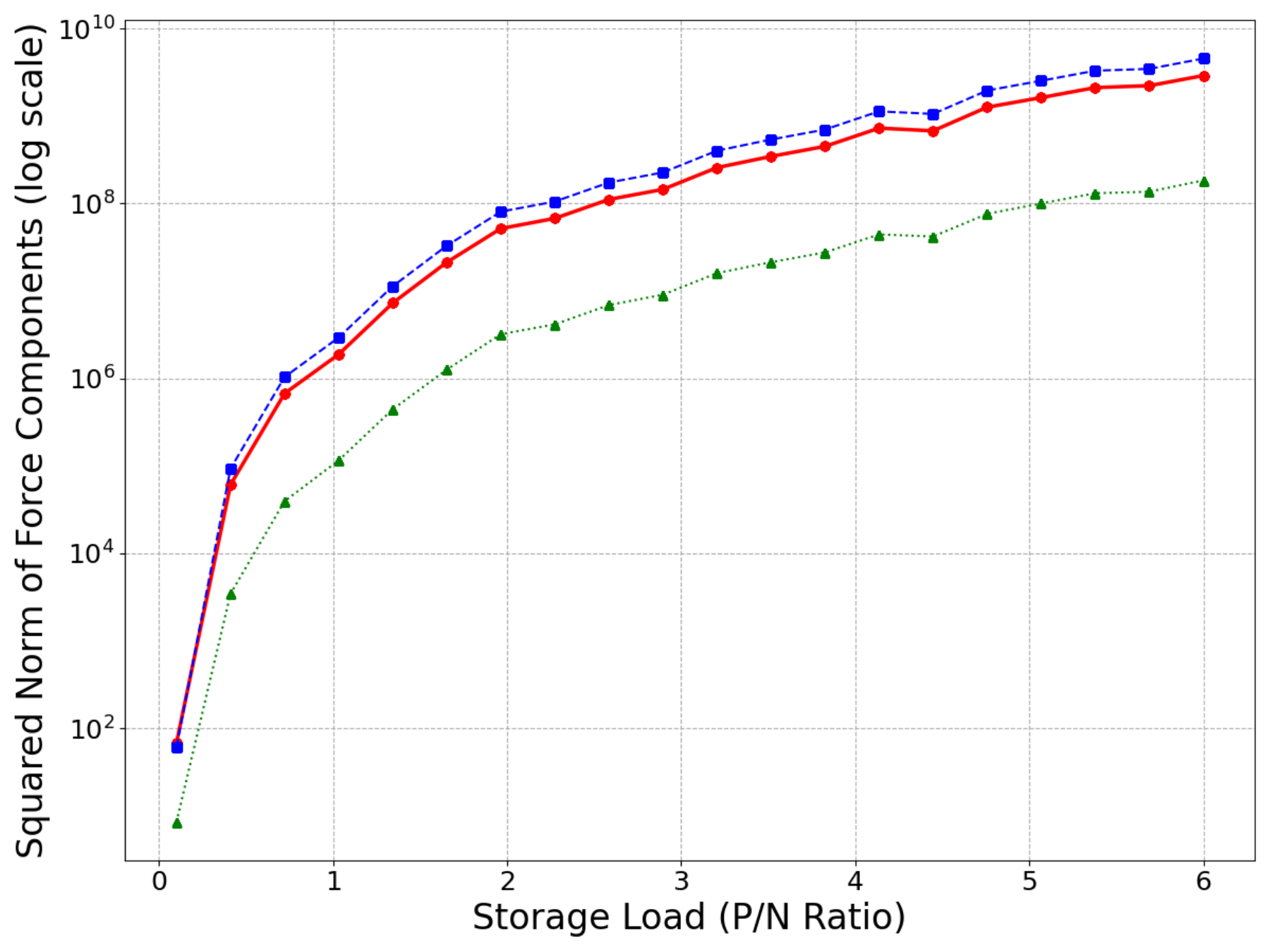}
  \caption{Growth of force component magnitudes as a function of
    storage load $P/N$ for a fixed small $\gamma$.  The plot shows the
    squared norms of the Pinnacle Sharpness $\|\nabla V\|^2$
    (\textbf{red solid line with circles}), the Direct Force
    $\|\bm{F}_d\|^2$ (\textbf{blue dashed line with squares}), and the
    Indirect Force $\|\bm{F}_i\|^2$ (\textbf{green dotted line with
      triangles}). The Direct Force grows significantly faster than
    the Indirect Force. Theoretically, this reflects the amplification
    of the leading spectral mode ($\lambda_1$) due to Spectral
    Concentration, which allows the driving force to overwhelm the
    interference and create a deep attractor basin.  }
  \label{fig:force_profile}
\end{center}
\end{figure}

The result is shown in Figure~\ref{fig:force_profile}. This plot
clearly reveals the mechanism behind the optimization ridge. As $P/N$
increases, both $\|\bm{F}_d\|^2$ (blue line) and $\|\bm{F}_i\|^2$ (green line)
increase by orders of magnitude. However, the direct force $\bm{F}_d$ grows
at a substantially faster rate than the indirect force $\bm{F}_i$. Because
the two forces are in strong opposition ($\rho \approx -1$), the total
Pinnacle Sharpness (red line) is determined by their difference,
$(\|\bm{F}_d\| - \|\bm{F}_i\|)^2$, and is therefore dominated by the explosive
growth of the direct force.

This leads to our final conclusion. The Ridge of Optimization is a
regime where: (1) the direct and indirect forces are in a state of
strong, cooperative anti-correlation, and (2) the learning algorithm,
driven by the high storage load, amplifies the magnitude of the direct
force $\bm{F}_d$ to a level that vastly overwhelms the opposing
feedback force $\bm{F}_i$. The resulting large net force creates the
exceptionally sharp and stable attractors observed. This is a
remarkable demonstration of a system achieving extreme stability not
through the synergy of forces, but through a controlled and amplified
antagonism.

\section{Theoretical Analysis: Spectral Mechanism of Self-Organization}
\label{sec:theory}

This section provides a theoretical interpretation of the
phenomenology reported in Section~\ref{sec:results}.

The phenomenological experiments in Section~\ref{sec:results}
presented a compelling puzzle: the emergence of the Ridge where
stability is maximized amidst strong Force Antagonism. In this
section, we solve this puzzle. We posit that these phenomena are
manifestations of a spectral phase transition in the learned
network. We first deduce this mechanism theoretically and then
validate it with targeted experiments.

\subsection{Theory: Spectral Mechanism of Stability}
\label{sec:spectral_theory}
To provide a theoretical explanation for the observed Ridge and the
associated Force Antagonism, we analyze the spectral properties of the
network in the feature space. We propose a two-step mechanism: first,
we link the geometric stability to the spectral properties of the
kernel, and second, we characterize the unique spectral state that
emerges on the Ridge.

\subsubsection{Step 1: The Geometry-Spectrum Link}
The local stability of an attractor is geometrically quantified by the
Pinnacle Sharpness $M$. Although the dynamics are defined by the
kernel update rule, the underlying curvature of the energy landscape
is governed by the Hessian matrix. By analyzing the Hessian in the
reproducing kernel Hilbert space (see Appendix A.2 for derivation), we
establish a direct link between the attractor's geometry and the
spectrum of the learned model.

\begin{proposition}[Stability-Spectrum Link]
  The Pinnacle Sharpness $M(\boldsymbol{\xi}^\mu)$ scales
  proportionally with the largest eigenvalue $\lambda_{\max}$ of the
  effective Gram matrix
  $\bm{K}_{\alpha} = \boldsymbol{\alpha}^{1/2} \bm{K}
  \boldsymbol{\alpha}^{1/2}$:
  \begin{equation}
    M(\boldsymbol{\xi}^\mu) \propto \lambda_{\max}(\bm{K}_{\alpha}),
  \end{equation}
  where the proportionality holds up to kernel-dependent constants under
  the local isometry approximation.
\end{proposition}
This proposition provides the fundamental principle: to maximize
attractor stability (and thus noise robustness), the learning
algorithm must find a configuration of the effective matrix
$\boldsymbol{\alpha}$ that maximizes the dominant spectral mode of the
system.

\subsubsection{Step 2: Spectral Concentration on the Ridge}
Based on Proposition 1, we can now interpret the Ridge. We hypothesize
that the high stability observed in this regime is not due to a
uniform strengthening of all weights, but rather due to a specific
reorganization of the spectral structure. We term this phenomenon
\textbf{Spectral Concentration}.

\begin{proposition}[Spectral Concentration, empirical and heuristic]
  On the Ridge of Optimization, the matrix $\boldsymbol{\alpha}$
  exhibits strong spectral concentration. The leading eigenvalue
  becomes dominant ($\lambda_1 \gg \lambda_k$ for $k > 1$), thereby
  maximizing the Pinnacle Sharpness via Proposition 1. Crucially,
  unlike a rank-1 collapse, the trailing eigenvalues remain non-zero
  ($\lambda_k > 0$), maintaining the full-rank structure necessary to
  store multiple patterns.
\end{proposition}

Appendix A.3 provides a theoretical justification for this
phenomenon. It explains how the maximum-margin objective naturally
leads to a spectral structure that amplifies the principal mode for
stability while preserving separability for capacity.

This spectral structure explains the dual nature of the network's high
performance:
\begin{enumerate}
   \item \textbf{Global Stability:} The massive leading eigenvalue
  $\lambda_1$ creates a steep, deep potential funnel (Direct Force)
  that strongly pulls states towards the stored patterns.
    \item \textbf{Memory Capacity:} The non-zero trailing eigenvalues
      preserve the fine-grained information required to distinguish
      between different patterns, preventing them from collapsing into
      a single spurious attractor.
\end{enumerate}

The Force Antagonism observed in Section~\ref{sec:results} is the
dynamical signature of this spectral hierarchy: the dominant mode
exerts the primary restorative force, while the subordinate modes
manage the interference to ensure separability.

One plausible mechanism underlying this spectral reorganization is
the interaction between margin maximization and regularization
inherent in kernel logistic regression, which jointly promotes
variance concentration along dominant directions while preserving
sufficient spectral diversity for class separability.  Although a
complete dynamical theory remains an open problem, this
interpretation is consistent with the empirical spectral trends
observed across all tested regimes.

\subsection{Empirical Validation of the Theory}
To confirm that Spectral Concentration is indeed the mechanism behind
the Ridge, we analyzed the spectral properties of the learned
effective matrix $\boldsymbol{\alpha} \in \mathbb{R}^{P \times P}$.

\begin{figure}[t]
    \centering
    \includegraphics[width=0.7\hsize]{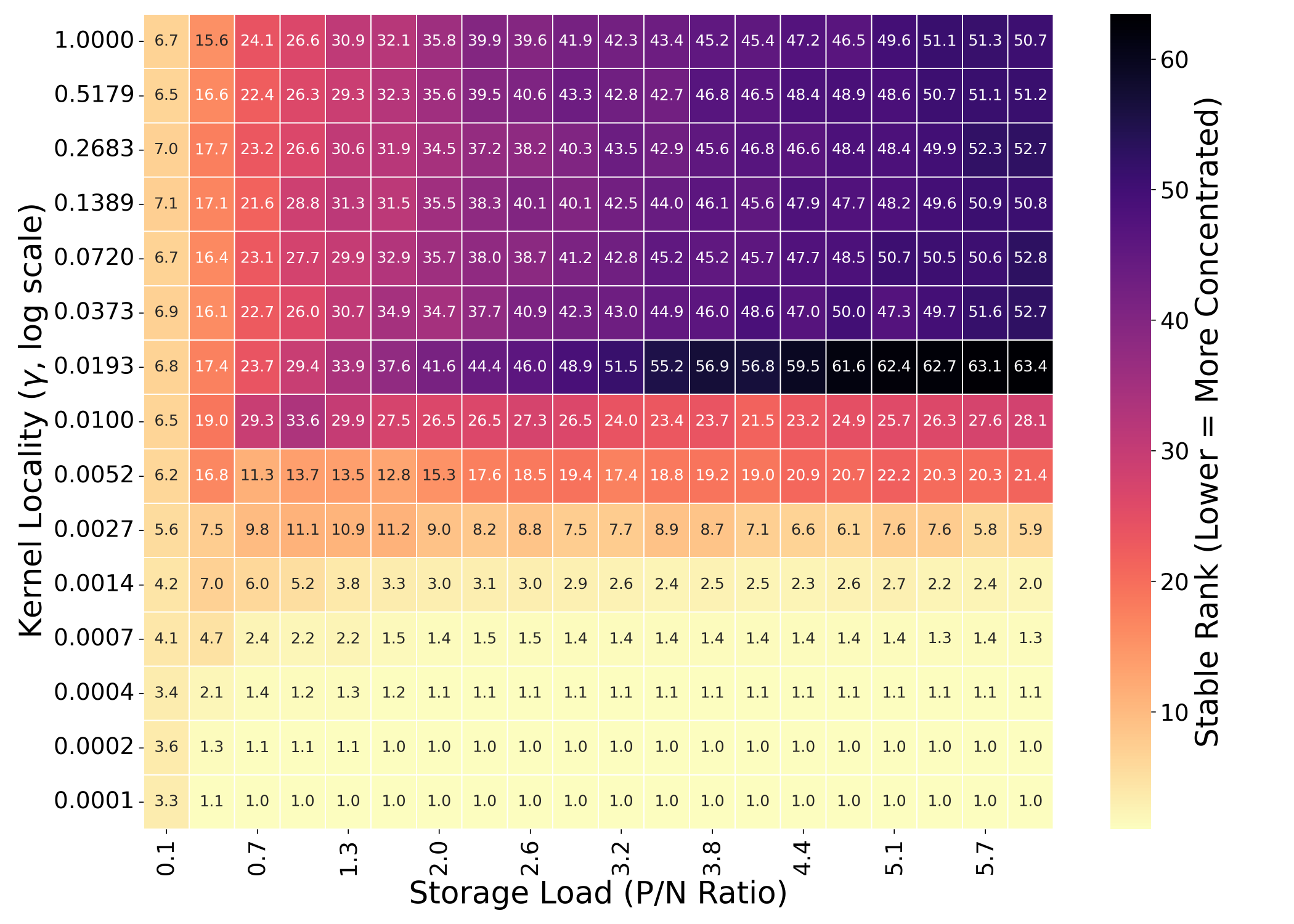}
    \caption{Phase diagram of Spectral Concentration, visualized as a
      heatmap of the Stable Rank.  The heatmap shows the Stable Rank
      (effective rank) of the effective pattern interaction matrix
      $\boldsymbol{\alpha}$ across the phase diagram.  Lower values
      (lighter/yellow colors) indicate a more concentrated spectrum.
      Crucially, the region of minimal Stable Rank perfectly aligns
      with the Ridge of Optimization observed in
      Fig.~\ref{fig:sharpness_diagram}, confirming that maximal
      attractor stability is achieved when the weight spectrum becomes
      highly concentrated (but not fully collapsed to rank 1).}
    \label{fig:stable_rank}
\end{figure}

\textbf{Effective Rank Analysis:} We utilized the \textit{Stable
  Rank}, also known as the numerical rank in the random matrix
literature~\cite{Rudelson2007}, to quantify the spectral distribution
of $\boldsymbol{\alpha}$. For singular values
$\sigma_1 \ge \sigma_2 \ge \dots \ge \sigma_P \ge 0$, the Stable Rank
is defined as:
\begin{equation}
  \text{StableRank}(\boldsymbol{\alpha}) \coloneqq \frac{\sum_{i=1}^P \sigma_i^2}{\sigma_1^2}.
  \label{eq:stable_rank}
\end{equation}
A lower Stable Rank indicates a more concentrated spectrum, where the
leading mode dominates the energy.  As shown in
Fig.~\ref{fig:stable_rank}, the region where the Stable Rank is
minimal perfectly overlaps with the Ridge observed in
Fig.~\ref{fig:sharpness_diagram}. Note that while the rank is
significantly reduced compared to the random initialization, it does
not collapse to exactly 1.0, consistent with the requirement for
multi-pattern storage.

\begin{figure}[t]
    \centering
    \includegraphics[width=0.65\hsize]{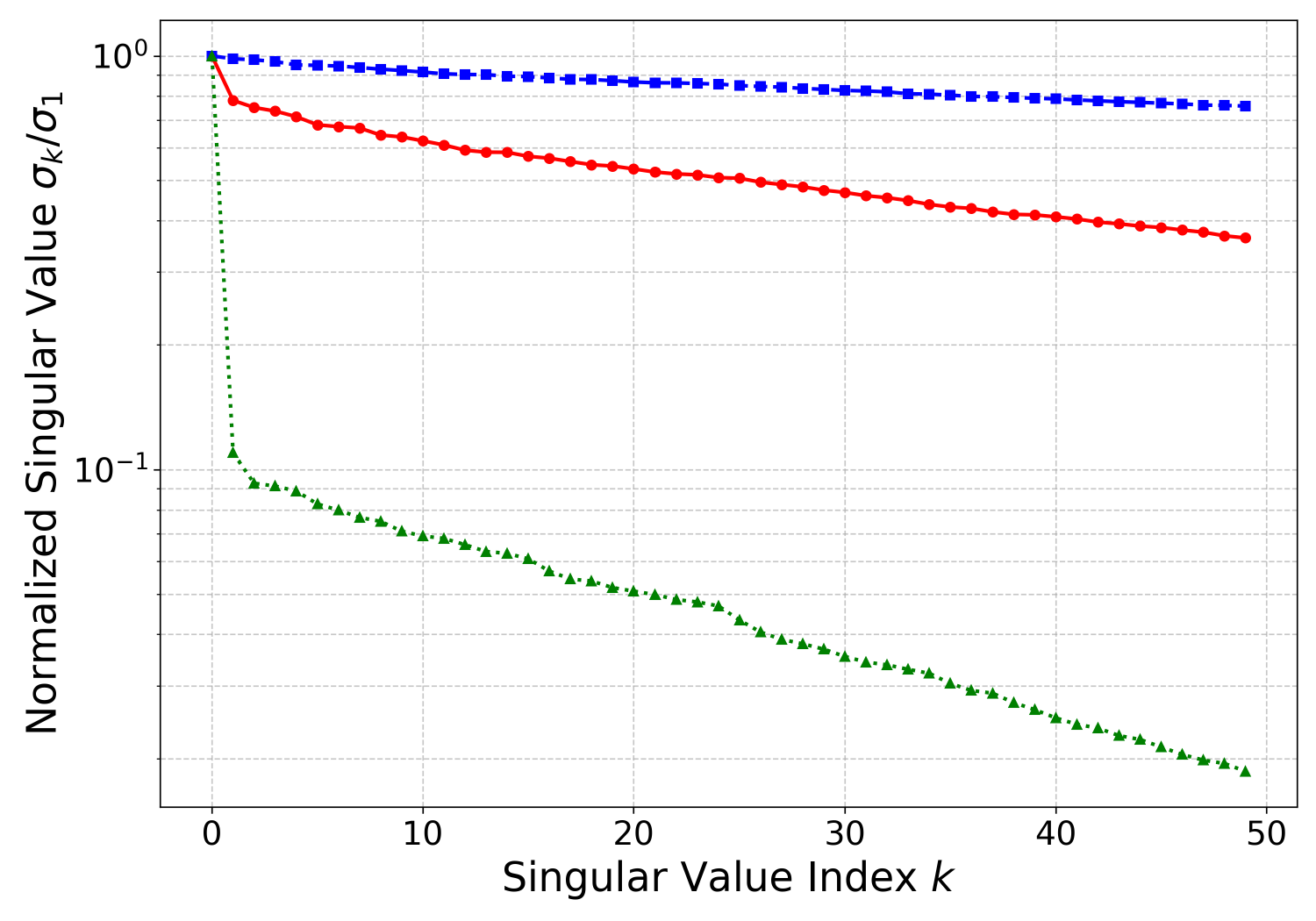}
    \caption{Spectral Concentration Analysis, showing a comparison of
      eigenvalue spectra.  The plot compares the eigenvalue spectra of
      the effective pattern interaction matrix $\boldsymbol{\alpha}$
      at three characteristic points in the phase diagram.  The
      \textbf{green dotted line (Global/Low Load)} exhibits a sharp
      rank-1 collapse ($\lambda_1 \gg \lambda_2 \approx 0$), leading
      to memory loss.  The \textbf{blue dashed line (Local/High
        Gamma)} shows a diffuse spectrum
      ($\lambda_k \approx \text{const}$), resulting in weak stability.
      The \textbf{red solid line (On Ridge)} demonstrates
      \textbf{Spectral Concentration}, characterized by a dominant
      leading eigenvalue for stability ($\lambda_1$) coexisting with a
      non-zero heavy tail ($\lambda_{k>1}$) for capacity. This
      ``L-shaped'' structure confirms the theoretical prediction of a
      critical balance.  }
    \label{fig:spectra}
\end{figure}

\textbf{Eigenvalue Spectrum:} We further examined the detailed
eigenvalue distribution of $\boldsymbol{\alpha}$ at specific points.
Fig.~\ref{fig:spectra} reveals a striking contrast that explains the
mechanism:
\begin{itemize}
\item \textbf{Off Ridge (Global):} The spectrum (green dotted line)
  shows a sharp collapse, resembling a rank-1 structure. This
  corresponds to the low-capacity regime where memories merge into a
  single attractor.
    \item \textbf{Off Ridge (Local):} The spectrum (blue dashed line)
      is flat and diffuse, indicating that the network fails to
      establish a dominant stability mode.
    \item \textbf{On the Ridge:} The spectrum (red solid line)
      exhibits a unique \textbf{intermediate structure}. It features a
      distinct leading mode ($\sigma_1$) that drives stability, yet
      crucially maintains a slowly decaying, heavy tail
      ($\sigma_{k>1}$). This confirms that the Ridge represents a
      Goldilocks zone where the network maximizes stability via
      spectral concentration without suffering from rank collapse.
\end{itemize}

These results provide direct empirical evidence that the Ridge
corresponds to a phase where the network self-organizes into a state
of extreme spectral concentration.

\subsection{Unified Interpretation}
The combination of phenomenological observation (Force Antagonism) and
spectral evidence (Spectral Concentration) leads to a unified physical
picture.  The Ridge represents a sweet spot where the kernel locality
$\gamma$ is tuned such that the system can amplify the principal
spectral mode to form deep, sharp attractors (via Proposition 1),
while effectively managing inter-pattern interference using the
remaining spectral degrees of freedom.  The Force Antagonism can be
interpreted as the dynamical manifestation of this spectral disparity:
the massive Direct Force arises from the dominant $\lambda_1$
component, while the Indirect Force reflects the collective
contribution of the trailing modes, acting to refine the basin
boundaries.

\section{Discussion}
\label{sec:discussion}

In this work, we have moved beyond simple performance metrics to
provide a clear physical picture of the self-organization mechanism in
KLR-trained Hopfield networks. Our central finding is the discovery of
a Ridge of Optimization and its spectral origin, Spectral
Concentration. Here, we discuss the broader implications of these
findings.

\subsection{The Ridge as a Critical Phase Boundary}
The spectral analysis in Fig.~\ref{fig:spectra} revealed that the
Ridge represents a critical state. The network does not simply
maximize the leading eigenvalue to the point of rank-1 collapse, which
would imply a trivial memory with unit capacity. Instead, it maintains
a delicate balance: maximizing $\lambda_1$ for global stability while
preserving a non-zero tail ($\lambda_{k>1}$) for pattern
discriminability.  This suggests that the Ridge can be interpreted as
a \textbf{phase boundary} or a Goldilocks zone between an ordered
phase (rank-1, low capacity) and a disordered phase (diffuse spectrum,
weak stability). This observation aligns with the hypothesis of
\textit{Self-Organized Criticality} in biological neural
networks~\cite{Beggs2003}, where systems tune themselves to a critical
point to maximize information processing capabilities.

It is important to distinguish Spectral Concentration from simple
dimensionality reduction techniques like Principal Component Analysis
(PCA). PCA blindly discards trailing eigenvalues to minimize
reconstruction error, often leading to the loss of pattern identity
(similar to rank-1 collapse). In contrast, KLR learning is
discriminative; it actively preserves the ``heavy tail'' of the
spectrum exactly to the extent required to maintain the decision
boundaries between patterns. Thus, the Ridge represents an optimal
compression that balances energy minimization (via the leading mode)
with classification margin (via the trailing modes).

\subsection{The Information Geometry of Learning}
Our spectral theory explains the static properties of the Ridge, but
how the learning dynamics navigate to this critical state remains an
open question.  Here, we provide an \emph{information-geometric
interpretation} of the empirically observed learning dynamics, which
suggests that the Ridge acts as a \textbf{geometric barrier} at the
brink of singularity.  As shown in Fig.~B-1, the learning dynamics at
the edge of the Ridge ($\gamma = 0.01$) exhibit a pronounced divergence
in the norm of the Natural Gradient.  This behavior is consistent with
an effective pinning of the learning trajectory near the boundary of
the manifold.

Importantly, the approach to this regime is not chaotic but highly
structured.  As illustrated in Fig.~B-2, the dynamics on the stable
part of the Ridge ($\gamma = 0.02$) display a two-phase learning
process: an initial phase dominated by the acquisition of global
stability, followed by a prolonged fine-tuning phase that refines
memory capacity.  From an information-geometric viewpoint, this
behavior can be interpreted as learning guided by an emergent curved
geometry.  Consistent with this interpretation, the learning
trajectory traces a distinct parabolic arc in parameter space
(Fig.~B-3), which is characteristic of gradient flow on a curved
statistical manifold.  For clarity, reduced basin stability occurs
primarily in the high-load region (large $\beta$), where interference
between stored patterns becomes significant.

\subsection{Comparison with Modern Hopfield Networks}
Recent advances in associative memory, such as Modern Hopfield
Networks (MHNs)~\cite{Krotov2016, Ramsauer2021}, achieve high capacity
by modifying the energy function to include steep non-linearities
(e.g., exponential or polynomial terms). In contrast, our approach
demonstrates that similar high-performance regimes can be accessed
within the standard quadratic energy framework (in feature space),
provided the weights are learned via a discriminative objective like
KLR.  This is significant because quadratic energy models are often
more amenable to implementation on neuromorphic hardware and
analytical tractability. Our results show that the \textit{learning
  rule} can be just as powerful as the \textit{architecture design} in
determining the memory landscape.

\subsection{Practical Implications and Future Directions}
The clean, spurious-free landscape achieved by Spectral Concentration
makes KLR Hopfield networks an attractive candidate for reliable
content-addressable memory systems. However, the computational cost of
kernel evaluations scales with the number of stored patterns. Future
work should explore sparse approximations (e.g., Nystr\"{o}m
method~\cite{Williams2000} or Random Fourier
Features~\cite{Rahimi2007}) to reduce the effective rank explicitly,
potentially enforcing the ``Spectral Concentration'' structure by
design to achieve linear scaling. Furthermore, extending this spectral
theory to analyze the dynamics of Transformer attention layers, which
share mathematical similarities with kernel memories, remains an
exciting avenue for research.

Furthermore, our discovery of Spectral Concentration provides a
theoretical justification for using low-rank approximations in kernel
machines. The fact that the effective weight matrix naturally exhibits
a low stable rank on the Ridge implies that sparse approximation
methods, such as the Nystr\"{o}m method, can
capture the essential dynamics without significant loss of
performance, paving the way for scalable implementations.

An important direction for future work is to examine
how the observed attractor statistics and scaling behavior depend on
modeling assumptions. In particular, alternative learning rules,
structured or sparse network topologies, and nonrandom data
distributions may alter the phase structure. A systematic
investigation of such variants would clarify the robustness and
generality of the phenomena reported here.

While our experiments focused on random pattern
ensembles to isolate spectral and interference effects, structured
or correlated datasets may alter the phase boundaries or reshape the
spectrum. Understanding how such statistical structure influences
the emergent spectral organization remains an important direction
for future investigation.

\section{Conclusion}
\label{sec:conclusion}
In this paper, we have presented a coherent theoretical account of
attractor self-organization in high-capacity kernel Hopfield networks.
Moving beyond conventional performance metrics, we investigated the
dynamical and spectral mechanisms that allow these networks to operate
near classical storage limits.

Our phenomenological analysis identified a Ridge of Optimization in
the phase diagram, where the network achieves maximal stability and
exhibits a distinctive Force Antagonism. We theoretically interpreted
this behavior through its spectral origin, which we termed
\textbf{Spectral Concentration}. On the Ridge, learning drives the
weight spectrum toward a critical regime in which the leading
eigenvalue is amplified to enhance global stability, while the
remaining eigenvalues remain finite, enabling high-capacity storage.
This picture goes beyond a simple rank-1 collapse and instead
characterizes the optimal memory landscape as a finely balanced
interplay between attractive forces and pattern discriminability.

By linking the geometry of attractors with the spectral structure of
the kernel machine, this work provides a principled framework for
understanding how learning induces self-organized memory states.
More broadly, the notion of spectral concentration offers insight into
how stability and capacity can be reconciled in associative systems,
and may inform the analysis of related learning dynamics in modern
kernel- and attention-based architectures.

\funding
Not applicable.

\conflictsofinterest
The author declares no competing interests.

\authorcontribution
The sole author contributed to the present work.

\aitools
The author used ChatGPT (GPT-5.2) and Gemini 2.5 Pro for proofreading
the English manuscript.

\appendix
\section{Spectral Theory of Kernel Attractors}
\label{app:spectral_theory}

This appendix provides the theoretical derivations linking the
attractor stability to the spectral properties of the kernel Gram
matrix, supporting the propositions in Section~\ref{sec:theory}.

\subsection{Notation and Setup}
Let $\mathcal{H}$ be the reproducing kernels Hilbert space associated
with the kernel $K(\bm{x}, \bm{y})$. The stored patterns are mapped to
feature vectors $\phi(\boldsymbol{\xi}^\mu) \in \mathcal{H}$.  We
define the feature matrix
$\boldsymbol{\Phi} = [\phi(\boldsymbol{\xi}^1), \dots,
\phi(\boldsymbol{\xi}^P)]$ and the Gram matrix
$\bm{K} = \boldsymbol{\Phi}^\top \boldsymbol{\Phi} \in \mathbb{R}^{P
  \times P}$.  The weight operator $\bm{W}$ in $\mathcal{H}$ is
represented as
$\bm{W} = \boldsymbol{\Phi} \boldsymbol{\alpha}
\boldsymbol{\Phi}^\top$, where
$\boldsymbol{\alpha} \in \mathbb{R}^{P \times P}$ is the effective
pattern interaction matrix, representing the collective learned
interactions.

\subsection{Proof of Proposition 1 (Stability-Spectrum Link)}
\label{app:proof_prop1}

We aim to show that the Hessian of the energy function, which governs
the local stability in the input space, is spectrally related to the
kernel Gram matrix in the feature space.

The energy function is defined as
$E(\bm{s}) = -\frac{1}{2} \langle \phi(\bm{s}), \bm{W} \phi(\bm{s})
\rangle$. Let $\bm{s}^*$ be a fixed point. The Hessian matrix
$H = \nabla^2 E(\bm{s}^*)$ is dominated by the first-order term
(Gauss-Newton approximation):
\begin{equation}
    H \approx J^\top \bm{W} J,
\end{equation}
where we define the sign convention such that $H$ represents the
magnitude of the curvature (positive definite around a stable
attractor), ignoring the negative sign from the energy definition.
$J$ is the Jacobian matrix of the feature map $\phi(\bm{s})$ at
$\bm{s}^*$.

To analyze the spectrum of $H$, consider the quadratic form with
respect to an arbitrary unit vector $\bm{u}$ in the input space
($\|\bm{u}\|^2=1$):
\begin{equation}
    \bm{u}^\top H \bm{u} = \bm{u}^\top (J^\top \bm{W} J) \bm{u} = (J \bm{u})^\top \bm{W} (J \bm{u}).
\end{equation}
Let $\bm{v} = J \bm{u}$ be the mapped vector in the feature space. A
key property of the RBF kernel is its local isometry; the mapping
scales local distances uniformly. Specifically, using the Taylor
expansion of the kernel, one can show that the squared norm scales as
$\|\bm{v}\|^2 = \bm{u}^\top (J^\top J) \bm{u} \approx 2\gamma
\|\bm{u}\|^2 = 2\gamma$.

Thus, maximizing the curvature $\bm{u}^\top H \bm{u}$ in the input
space is equivalent to maximizing the quadratic form
$\bm{v}^\top \bm{W} \bm{v}$ in the feature space, subject to the norm
constraint on $\bm{v}$.  Substituting the weight operator
$\bm{W} = \boldsymbol{\Phi} \boldsymbol{\alpha}
\boldsymbol{\Phi}^\top$, the relevant spectral properties are
determined by the non-zero eigenvalues of the matrix product
$\boldsymbol{\Phi}^\top \bm{W} \boldsymbol{\Phi}$ (restricted to the
span of patterns), which reduces to analyzing the effective Gram
matrix:
\begin{equation}
    \bm{K}_{\alpha} = \boldsymbol{\alpha}^{1/2} \bm{K} \boldsymbol{\alpha}^{1/2}.
    \label{eq:effective_gram_app}
\end{equation}
Note that the non-zero eigenvalues of $\bm{K}_{\alpha}$ are identical
to those of $\boldsymbol{\alpha} \bm{K}$.  Consequently, the Pinnacle
Sharpness $M(\boldsymbol{\xi}^\mu)$, which corresponds to the squared
gradient norm (and scales with the dominant curvature), is
proportional to the leading eigenvalue of this matrix:
\begin{equation}
    M \propto \lambda_{\max}(\bm{K}_{\alpha}).
\end{equation}
This confirms Proposition 1.

\subsection{Theoretical Basis for Spectral Concentration (Proposition 2)}
\label{app:spectral_concentration}
Here we provide a heuristic justification for why the optimization
leads to Spectral Concentration on the Ridge.

In the high-load or high-regularization regime, Kernel Logistic
Regression approximates a maximum-margin objective in the feature
space. The dual optimization problem involves maximizing a quadratic
form $\boldsymbol{\alpha}^\top \bm{K} \boldsymbol{\alpha}$ subject
to constraints.  Let $\bm{v}_1$ be the eigenvector of the Gram
matrix $\bm{K}$ associated with the largest eigenvalue
$\lambda_1$. If there is a spectral gap ($\lambda_1 > \lambda_2$), the
direction $\bm{v}_1$ offers the most efficient way to increase the
margin (or stability).

Consequently, the optimization algorithm is biased to align the weight
vector $\boldsymbol{\alpha}$ primarily with $\bm{v}_1$ to maximize
the objective. This creates a large disparity in the spectrum of
$\boldsymbol{\alpha}$, where the component along $\bm{v}_1$ is
amplified:
\begin{equation}
  \frac{\bm{v}_1^\top \boldsymbol{\alpha} \bm{v}_1}{\bm{v}_k^\top \boldsymbol{\alpha} \bm{v}_k} \gg 1 \quad \text{for } k > 1.
\end{equation}
However, unlike a pure rank-1 collapse, the constraints of storing $P$
distinct patterns require that $\boldsymbol{\alpha}$ retains non-zero
components along other directions ($\bm{v}_{k>1}$) to ensure linear
separability of all patterns. This balance results in the observed
Spectral Concentration: a dominant $\lambda_1$ for stability,
coexisting with a non-zero tail for capacity.

\section{Information Geometric Formulation}
\label{app:info_geo}

This appendix provides a brief overview of the information geometric
concepts used in Section~\ref{sec:discussion} to analyze the learning
dynamics. We follow the standard formulation by
Amari~\cite{Amari2016}.

\subsection{Statistical Manifold and Fisher Metric}
The set of probability distributions defined by the KLR model forms a
statistical manifold $M$.  Each point on this manifold corresponds to
a specific set of parameters $\boldsymbol{\alpha}$, which in
information geometry are known as the \textit{natural parameters} (or
$\theta$-coordinates).  The distance between two nearby points on this
manifold, corresponding to parameters $\boldsymbol{\alpha}$ and
$\boldsymbol{\alpha} + d\boldsymbol{\alpha}$, is measured by the
infinitesimal change in the probability distribution. This distance is
quantified by the Fisher Information Matrix (FIM)
$\bm{G}(\boldsymbol{\alpha})$ , which acts as a Riemannian metric
tensor.

For a single neuron, the probability of the output being $+1$ is given
by the sigmoid of the logit
$u = \sum_{u}\alpha_{u} K(\cdot, \boldsymbol{\xi}^{\mu})$.  This follows
a Bernoulli distribution, which is a member of the exponential
family. For such models, the FIM is defined as the expectation of the
outer product of the gradient of the log-likelihood:
\begin{equation}
G_{\mu\nu}(\boldsymbol{\alpha}) = \mathbb{E} \left[ \frac{\partial \log p}{\partial \alpha_\mu} \frac{\partial \log p}{\partial \alpha_\nu} \right].
\end{equation}
A key property of the logistic function is that
$\partial p/\partial \alpha_{\mu} = (y-p)K(\cdot, \boldsymbol{\xi}^{\mu})$,
where $y$ is the target and $p$ is the model's output
probability. Taking the expectation over the data distribution (and
assuming independence), this leads to the specific form used in the
main text:
\begin{equation}
\bm{G} = \bm{K} \bm{D} \bm{K},
\end{equation}
where $\bm{D}$ is a diagonal matrix with entries
$D_{\mu\mu} = p_{\mu}(1 - p_{\mu})$ representing the prediction
variance for pattern $\boldsymbol{\xi}^{\mu}$.

\subsection{Natural Gradient Descent}
Standard gradient descent (SGD) follows the steepest descent direction
in the Euclidean parameter space. However, this path is not
necessarily the steepest on the curved statistical manifold. The
direction of steepest descent on the manifold is given by the
\textit{Natural Gradient}~\cite{Amari1998}.

The Natural Gradient $\tilde{\nabla}L$ is defined as the direction
that maximizes the decrease in the loss function $L$ for a small,
fixed ``information distance'' (measured by the KL-divergence).  For
an infinitesimal step, the KL-divergence between
$p(\boldsymbol{\alpha})$ and
$p(\boldsymbol{\alpha} + d\boldsymbol{\alpha})$ is approximated by the
quadratic form of the Fisher metric:
\begin{equation}
D_{\text{KL}}(p(\boldsymbol{\alpha}) || p(\boldsymbol{\alpha} + d\boldsymbol{\alpha})) \approx \frac{1}{2} d\boldsymbol{\alpha}^\top \bm{G}(\boldsymbol{\alpha}) d\boldsymbol{\alpha}.
\end{equation}
Minimizing the change in loss $\nabla L^{\top} d\boldsymbol{\alpha}$
subject to a fixed information distance leads to the Natural Gradient
direction:
\begin{equation}
  \tilde{\nabla} L(\boldsymbol{\alpha}) = \bm{G}(\boldsymbol{\alpha})^{-1} \nabla L(\boldsymbol{\alpha}).
\end{equation}
The Natural Gradient Descent (NGD) update rule is therefore
$\boldsymbol{\alpha}_{t+1} = \boldsymbol{\alpha}_{t} - \eta
\tilde{\nabla}L (\boldsymbol{\alpha}_{t})$. This formulation allows us
to analyze the learning dynamics not as a simple descent in Euclidean
space, but as a flow on a curved Riemannian manifold, revealing the
geometric distortions induced by the FIM.

\subsection{Experimental Analysis of Learning Dynamics}
To validate the geometric interpretations presented in the Discussion,
we conducted numerical analyses of the SGD learning trajectory. We
tracked the parameter vector $\boldsymbol{\alpha}_{t}$ for a single
representative neuron over 200 training steps under different
hyperparameter regimes. 

\begin{figure}[t]
\centering
\includegraphics[width=0.7\columnwidth]{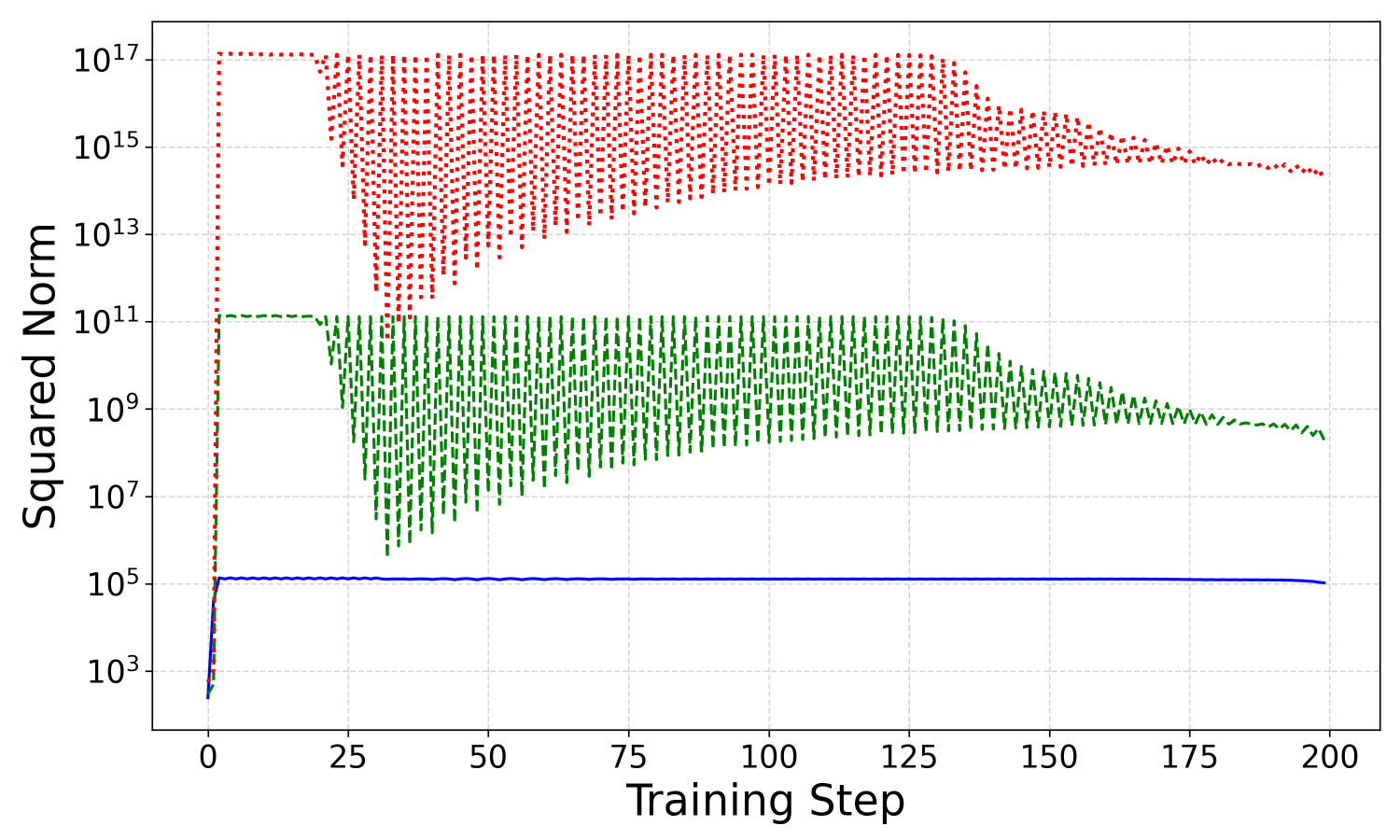}
\caption{Gradient Norm Dynamics on the Edge of the Ridge
  ($\gamma=0.01$).  The plot compares the squared norms of the
  Euclidean gradient $\|\nabla L\|^2$ (\textbf{blue solid line}), the
  Riemannian gradient $\|\nabla L\|^2_{G^{-1}}$ (\textbf{green dashed
    line}), and the Natural Step vector $\|G^{-1}\nabla L\|^2$
  (\textbf{red dotted line}).  The divergence of the Riemannian and
  Natural Step norms indicates that the manifold is approaching a
  singularity, creating a geometric barrier for SGD.  }
\label{fig:grad_norms}
\end{figure}

\textbf{Gradient Norm Dynamics:} We computed the squared norms of
three distinct gradient vectors at each step $t$: the Euclidean norm
$\|\nabla L(\boldsymbol{\alpha}_{t})\|^{2}$, the Riemannian norm
$\|\nabla L(\boldsymbol{\alpha}_{t}) \|^{2}_{G^{-1}} = \nabla L^{\top}
G^{-1} \nabla L$, and the Natural Step norm
$\|G^{-1}\nabla L(\boldsymbol{\alpha}_{t}) \|^{2}$.  As shown in
Fig.~\ref{fig:grad_norms}, the dynamics on the edge of the Ridge
($\gamma=0.01$) exhibit a divergence in the Riemannian and Natural
Step norms. This indicates that the manifold is approaching a
singularity, creating a geometric barrier for SGD.

\begin{figure}[t]
\centering
\includegraphics[width=0.7\columnwidth]{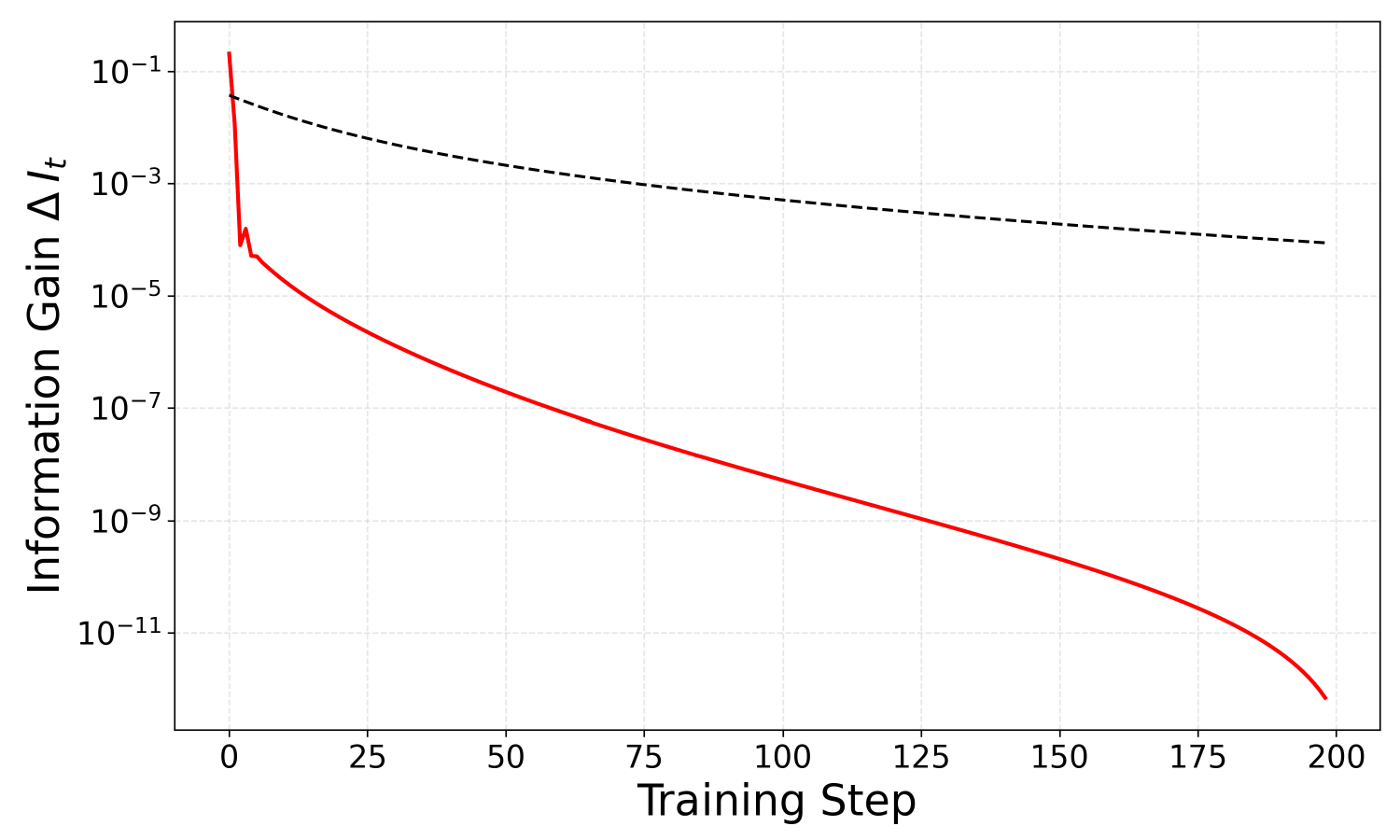}
\caption{Information Gain Decomposition on the Ridge ($\gamma=0.02$).
  The plot shows the information gain decomposed into the Principal
  Gain (\textbf{red solid line}) and the Tail Gain (\textbf{black
    dashed line}).  The dynamics exhibit a two-phase process: an
  initial phase dominated by Principal Gain is followed by a prolonged
  Fine-tuning phase dominated by Tail Gain.}
\label{fig:info_gain}
\end{figure}

\textbf{Pythagorean Decomposition of Information Gain:} Second, we
decomposed the information gain at each step,
$\Delta I_{t} \approx \frac{1}{2} \boldsymbol{\delta}_{t}^{T} G^{*}
\boldsymbol{\delta}_{t}$ into two orthogonal components based on the
eigenspace of the final FIM $G^{*}$: the Principal Gain (contribution
from $\lambda_1$ and the Tail Gain (contribution from
$\lambda_{k>1}$).  As shown in Fig.~\ref{fig:info_gain}, the dynamics
on the stable part of the Ridge ($\gamma=0.02$) exhibit a two-phase process: an
initial phase dominated by Principal Gain is followed by a prolonged
Fine-tuning phase dominated by Tail Gain, demonstrating a hierarchical
learning strategy.

\begin{figure}[h]
\centering
\includegraphics[width=0.7\columnwidth]{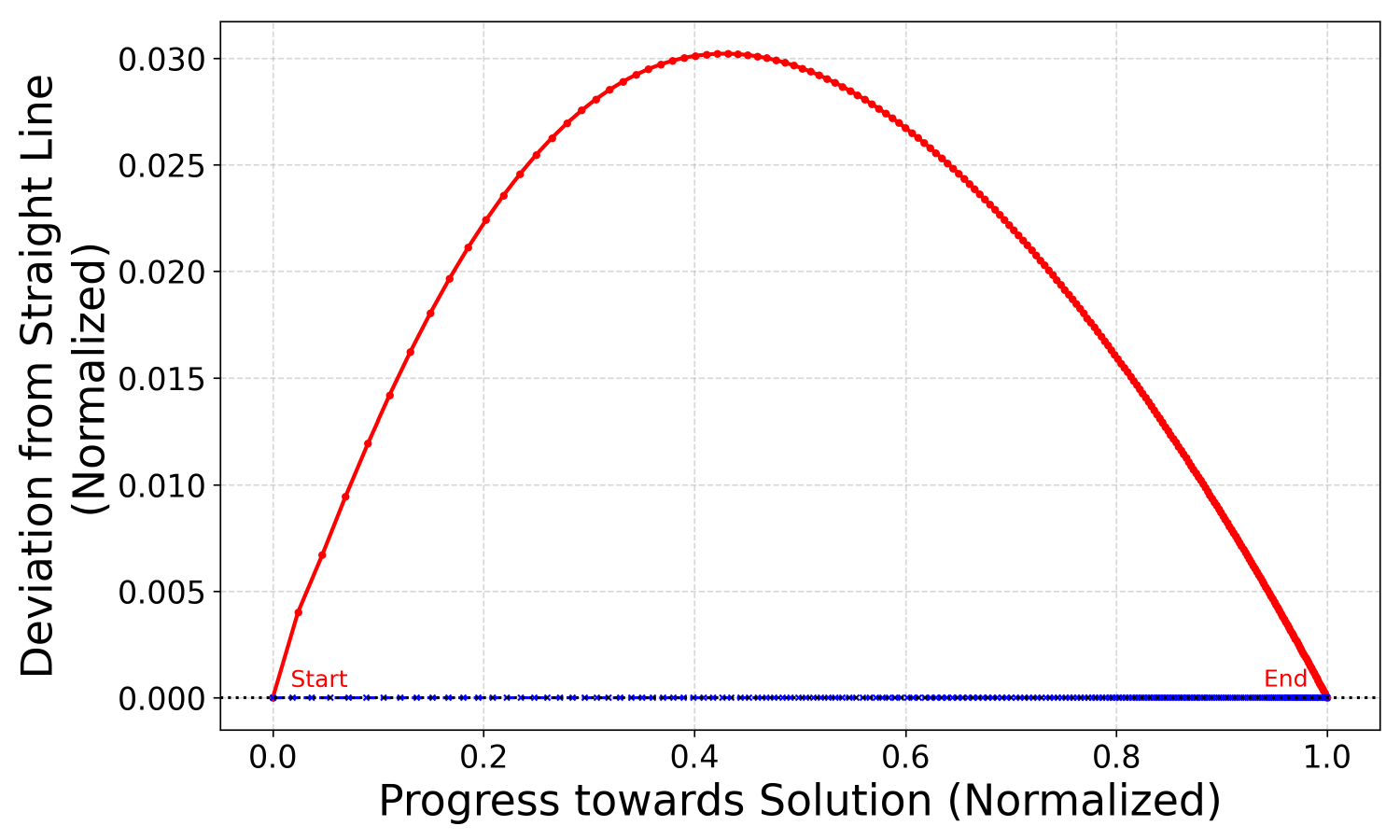}
\caption{Geodesic Deviation of Learning Trajectories.  The plot
  compares the deviation from the ideal e-geodesic (\textbf{black
    dotted line}) for the Ridge trajectory ($\gamma=0.02$, \textbf{red
    solid line}) and the Local trajectory ( $\gamma=0.1$, \textbf{blue
    dashed line}).  The Ridge trajectory's parabolic arc demonstrates
  that it is guided by the non-trivial manifold curvature, whereas the
  Local trajectory's straight path indicates a flat learning
  landscape.  }
\label{fig:geodesic_dev}
\end{figure}

\textbf{Geodesic Deviation:} Finally, we analyzed the shape of the
trajectory $\boldsymbol{\alpha}_{t}$ by plotting its deviation from
the e-geodesic (straight line) against its progress towards the final
state $\boldsymbol{\alpha}^{*}$.  As shown in
Fig.~\ref{fig:geodesic_dev}, the results reveal a stark contrast
between the two regimes: while the Local trajectory ($\gamma=0.1$) is
perfectly straight, the Ridge trajectory ($\gamma=0.02$) forms a
distinct parabolic arc. This confirms that the trajectory on the Ridge
is guided by the non-trivial curvature of the statistical manifold.

\end{document}